%% file: custom.tex
\pdfoutput=1

\documentclass[11pt]{article}

\usepackage[colorlinks = true,
            linkcolor = blue,
            urlcolor  = blue,
            citecolor = blue,
            anchorcolor = blue]{hyperref}
\newcommand{\MYhref}[3][blue]{\href{#2}{\color{#1}{#3}}}%

\usepackage[preprint]{acl}

\usepackage{times}
\usepackage{latexsym}
\input{latex/99-macros}

\usepackage[T1]{fontenc}

\usepackage[utf8]{inputenc}

\usepackage{microtype}

\usepackage{inconsolata}

\usepackage{graphicx}
\usepackage{multirow}
\usepackage{amssymb}
\usepackage{float}
\usepackage{verbatim}
\usepackage{listings}
\usepackage{booktabs}
\usepackage{amsmath}
\usepackage{textgreek}
\usepackage{xcolor}
\usepackage{color, colortbl,booktabs}
\definecolor{graycolor}{gray}{0.9}
\definecolor{bluecolor}{RGB}{11,68,153}

\title{Can Stories Help LLMs Reason?

Curating Information Space Through Narrative
}

\author{Vahid Sadiri Javadi \textsuperscript{\textPsi} \hspace{0.3cm} Johanne R. Trippas \textsuperscript{\maltese} \hspace{0.3cm} Yash Kumar Lal \textsuperscript{\textpsi} \hspace{0.3cm} Lucie Flek \textsuperscript{\textPsi} \\ 
\textsuperscript{\textPsi} University of Bonn, \textsuperscript{\maltese} RMIT University, \textsuperscript{\textpsi} Stony Brook University \\
\texttt{\{\MYhref[black]{mailto:vahid.sadirij@uni-bonn.de}{vahid.sadirij}, \MYhref[black]{mailto:lflek@uni-bonn.de}{lflek}\}@uni-bonn.de}, \\
\texttt{\MYhref[black]{mailto:j.trippas@rmit.edu.au}{j.trippas@rmit.edu.au}, \MYhref[black]{mailto:ylal@cs.stonybrook.edu}{ylal@cs.stonybrook.edu}}}

\begin{document}
\maketitle
\begin{abstract}
Narratives are widely recognized as a powerful tool for structuring information and facilitating comprehension of complex ideas in various domains such as science communication. 
This paper investigates whether incorporating narrative elements can assist Large Language Models (LLMs) in solving complex problems more effectively. 
We propose a novel approach, \textbf{Story of Thought (SoT)}, integrating narrative structures into prompting techniques for problem solving.
This approach involves constructing narratives around problem statements and creating a framework to identify and organize relevant information.
%
Our experiments show that using various LLMs with SoT consistently surpasses using them with other techniques on physics, chemistry, math, and biology questions in both the GPQA and JEEBench datasets.
The narrative-based information curation process in SoT enhances problem comprehension by contextualizing critical in-domain information and highlighting causal relationships within the problem space. 




\end{abstract}

\input{latex/01-introduction}
\input{latex/02a-related-work}
\input{latex/03-methodology}
\input{latex/04-experimenal-setup}
\input{latex/05-results}

\input{latex/05a-analysis}
\input{latex/06-discussion}
\input{latex/07-conclusion}
\input{latex/08-limitations}
\bibliography{custom, anthology}
\input{latex/09-appendix}
\end{document}

%% file: latex/99-macros.tex
\usepackage{hyperref}
\usepackage[]{xcolor}
\usepackage{enumitem}
\usepackage{graphicx}
\usepackage{natbib}

\usepackage{xcolor}
\definecolor{dgrey}{gray}{0.2}

\newcommand{\jt}[1]{\textcolor{teal}{\textbf{[[JT: #1]]}}} 
\newcommand{\vs}[1]{\textcolor{purple}{\textbf{[[VS: #1]]}}} 

\newcommand{\todo}[1]{\textcolor{red}{[[#1]]}} 

%% file: latex/01-introduction.tex
\section{Introduction}
\label{sec:intro}
Humans have an exceptional ability to understand and reason through narratives. A narrative-driven approach can enhance the comprehension and retention of complex subjects compared to simple fact listing~\cite{fisher2021human, abbott2020cambridge, gottschall2012storytelling}. For example, storytelling effectively structures information in science communication~\cite{dahlstrom2014using, norris2005theoretical, martinez2017finding}, education~\cite{engel2018rethinking, negrete2004learning}, and health communication~\cite{dudley2023use}, revealing relationships and contextual nuances~\cite{zak2015inspiring}. 
While 
\textit{narrative approach} contextualizes facts within a daily life scenario (story) with a planned structure, 
a
 \textit{factual approach} 
conveys information in a concise in-domain manner.


To date, large language models (LLMs) struggle with complex problem-solving tasks that require the ability to integrate, structure, and apply relevant information effectively~\cite{qiao-etal-2023-reasoning, wang2023scibench}. Prompting techniques based on breaking tasks into smaller subtasks, such as Chain-of-Thought (CoT)~\cite{wei2022chain} and its more recent adaptations~\cite{xia2024beyond}, have led to considerable improvements in problem-solving benchmarks.
The strategies of constructing natural language rationales ~\cite{ling-etal-2017-program}, in the CoT context also called reasoning processes, play a vital role in LLM prompting~\cite{ye2022unreliability, min2022rethinking,Wang2022SelfConsistencyIC,li2023-making}.

Inspired by the effectiveness of narrative in~\textit{(i)} identifying and explaining abstract concepts and~\textit{(ii)} organizing the information flow coherently, we explore integrating narrative elements into prompt-driven reasoning.
The main research questions addressed in this work are:
\begin{itemize}[leftmargin=*,noitemsep,topsep=0pt]

    \item[] \textbf{RQ1:} Can LLMs generate coherent and relevant narratives around problem statements to facilitate comprehension and reasoning?

   \item[] \textbf{RQ2:} Can incorporating narrative elements into prompting techniques improve model performance on complex problem-solving tasks? 
\end{itemize}


We make the following contributions:
\textit{(i)} We introduce a novel method, \textbf{Story of Thought (SoT)}, that aids LLMs to identify and arrange relevant information for solving complex tasks by incorporating narrative structures into the prompting process, \textit{(ii)} We evaluate the effectiveness of SoT on GPQA and JEEBench datasets of complex problems, showing superior performance to existing prompting techniques with SotA models, and
\textit{(iii)} We analyze the impact of narrative techniques to generate narrative-based explanations and investigate why they improve LLMs' reasoning abilities. 

%% file: latex/02a-related-work.tex
\section{Related Work}
\label{sec:related-work}

\citet{bruner1991narrative} posit that narratives are a fundamental mode of human thought, allowing individuals to convey complex concepts in a more understandable manner.
Presenting information through narratives can enhance learning and memory, promote engagement and motivation \cite{willingham2004ask, chen2023design}.
The development of narrative-based educational strategies \cite{bower1969narrative, mawasi2020systematic, norris2005theoretical} paved the way for using them as a framework for organizing information for problem solving.
The use of narratives can break down complex problems into sub-problems, providing a step-by-step approach to answering a question \cite{szurmak2013tell}.
\citet{10.1145/3640794.3665884} use different narratives techniques to satisfy diverse requirements for conversational information-seeking systems.

There are a plethora of datasets focusing on answering questions about given contexts.
Reading comprehension datasets \cite{khashabi-etal-2018-looking, welbl-etal-2018-constructing, williams-etal-2018-broad, mihaylov-etal-2018-suit} explicitly evaluate a system’s ability to answer questions that need information from multiple sentences in a passage.
NarrativeQA \cite{kocisky-etal-2018-narrativeqa} provides a dataset of 1,567 narratives and associated QA pairs as written by human annotators.
ROCStories \cite{mostafazadeh-etal-2016-corpus} is a collection of 5 sentence short stories over which numerous datasets such as TellMeWhy \cite{lal-etal-2021-tellmewhy} have been built to facilitate answering questions about narratives.
However, none of these datasets use narratives as a tool of understanding, or relate to problem solving.

Problem solving datasets focus on mathematics, physics or other scientific domains.
GSM8K \cite{cobbe2021gsm8k} is a dataset of 8.5K high quality linguistically diverse grade school math word problems created by human problem writers.
SciQ \cite{welbl-etal-2017-crowdsourcing} is built using a novel method for obtaining high-quality, domain-targeted multiple choice questions from crowd workers, and contains 13.7K multiple choice science exam questions.
ScienceQA \cite{lu2022learn} adds multimodal context to collected elementary and high school science questions.
While there has been rapid progress on these tasks, prior work has not integrated educational strategies such as narratives to tackle them, a setting which is likely to be used in the real world.
MedMCQA \cite{medmcqa} contains MCQ questions designed to address real-world medical entrance exam questions.
Such datasets have been used extensively as yardsticks to measure the progress of NLP techniques.

The strength of modern LLMs, coupled with the paradigm of prompting, has driven up performance on problem solving tasks.
In-context learning through few-shot examples has been used to teach LLMs about new tasks using a small number of examples.
Chain of thought prompting \cite{wei2022chain} nudges LLMs to generate intermediate steps to mimic an explicit reasoning process before answering a question.
Similarly, Tree of Thoughts (ToT) \cite{yao2023tree} and Graph of Thoughts (GoT) \cite{besta2024got} induce intermediate reasoning structures, trees and graphs respectively, to decide on an answer.
However, despite the fact that narratives have been used as a way to simplify problems, they have never been explored to improve the problem solving abilities of LLMs.

%% file: latex/03-methodology.tex
\begin{figure*}[hbt!]
\centering
\includegraphics[width=\textwidth]{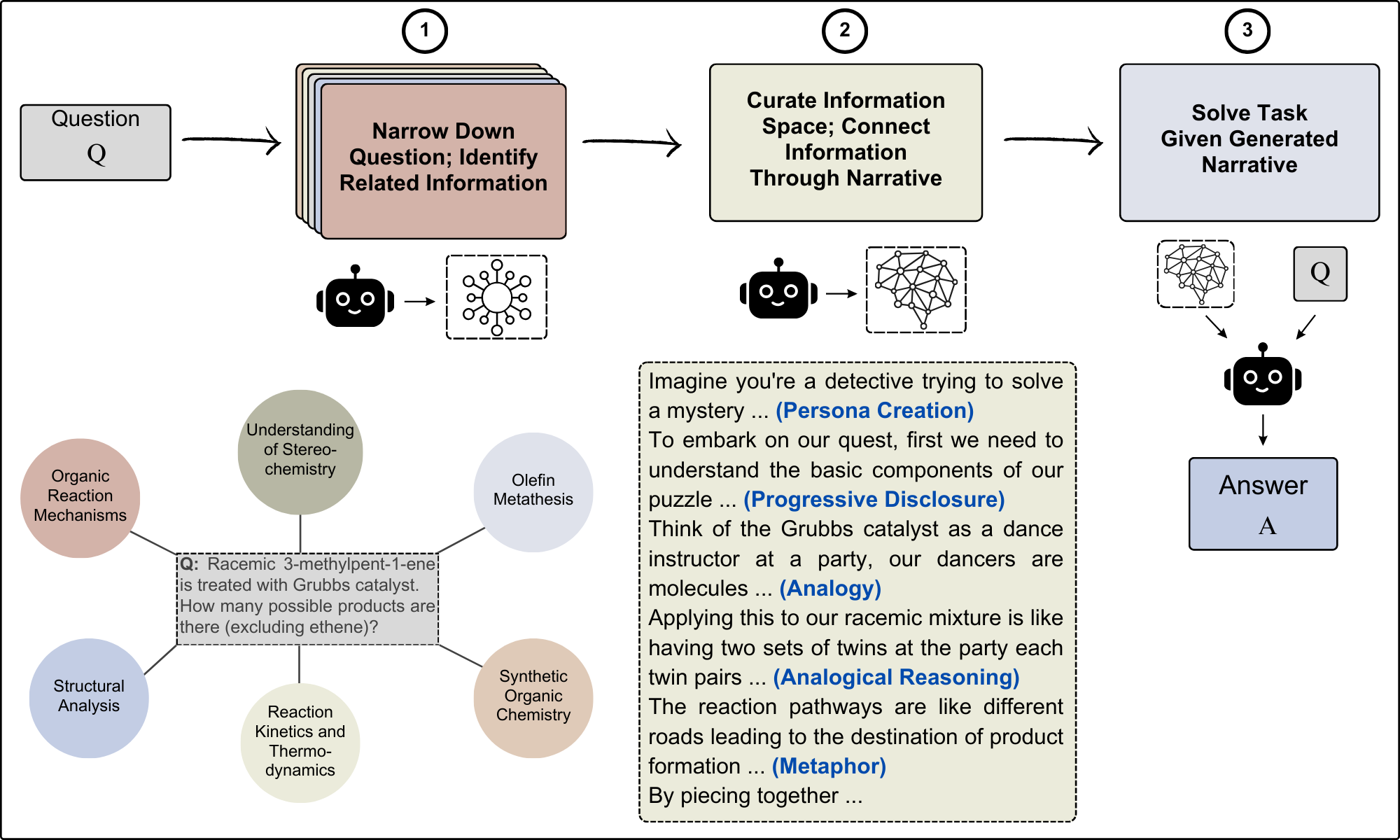}
\caption{A high-level overview of \textbf{Story of Thought} (SoT), consisting of three steps (top): \raisebox{.5pt}{\textcircled{\raisebox{-.9pt} {1}}}~Question Clarification,
\raisebox{.5pt}{\textcircled{\raisebox{-.9pt} {2}}}~Narrative Generation,
\raisebox{.5pt}{\textcircled{\raisebox{-.9pt} {3}}}~Solving Task and an actual example of LLM output (bottom) in each step for the GPQA task. The prompt designed for step 2 incorporates the narrative techniques (highlighted in blue) such as \textit{analogical reasoning}, which identifies similarities between the target concept (information being conveyed) and a more familiar concept (\textit{analogy}) and \textit{progressive disclosure} which reveals information gradually throughout the narrative, rather than presenting it all at once. See \autoref{appsec:prompts} for prompts for each step and \autoref{sot-cot-examples} for an example.}
\label{pipeline}
\end{figure*}

\section{Methodology: Story of Thought}
\label{sec:methodology}
We introduce \textbf{Story of Thought} (SoT), a novel prompt-driven reasoning approach that generates narrative-based clarification to guide LLMs' reasoning process. Inspired by 
the narrative format, 
the SoT approach leverages the cognitive benefits of storytelling, such as contextual understanding and relational reasoning, that can help LLMs identify and maintain the information structure.

Figure~\ref{pipeline} gives an overview of SoT. It involves three steps using narrative techniques: 
\textit{(i)}~\textbf{Question clarification} (i.e., acting as an explorer to dissect and clarify complex questions (Section~\ref{subsec:question-clarification})); 
\textit{(ii)}~\textbf{Narrative Generation} (i.e., generating detailed narratives from the clarified question components using different narrative techniques (Section~\ref{subsec:narrative-generation})); and 
\textit{(iii)}~\textbf{Problem Solving} (i.e., leveraging narratives to prompt the LLMs to solve the tasks (Section~\ref{subsec:solving-task})). 
We describe the exact prompts used in each step in \autoref{appsec:prompts}.

\subsection{Step 1: Question Clarification}
\label{subsec:question-clarification}
In the first step, 
we use the LLM's ability to explore and clarify the problem.
Starting with a specialized prompt, the LLM breaks down the question into its core components, identifying relevant subtopics and areas. This detailed analysis is crucial for generating a coherent narrative that thoroughly addresses the question. 

\subsection{Step 2: Narrative Generation}
\label{subsec:narrative-generation}
The second step involves generating detailed narratives based on the breakdown and clarification performed in Step 1 (question clarification). These narratives provide a structured context for the questions to enhance the LLM's understanding, reasoning, and problem-solving abilities. 
\citet{10.1145/3640794.3665884} discuss different narrative techniques required in conversational information seeking systems.
We integrate the below subset of these techniques into our prompt and task LLMs to generate a narrative, based on the information from Step 1:

\begin{enumerate}[leftmargin=*,noitemsep,topsep=0pt]
    \item \textbf{Progressive Disclosure}: Reveals information gradually, guiding the LLM step-by-step through the problem-solving process.
    \item \textbf{Branching}: Explores different paths or approaches to understanding the problem by providing multiple perspectives.
    \item \textbf{Analogy}: Uses comparisons to familiar concepts or situations to make abstract components more understandable.
    \item \textbf{Analogical Reasoning}: Facilitates understanding by reasoning through similarities between the problem and known situations.
    \item \textbf{Metaphor}: Simplifies complex ideas through metaphorical representation.
\end{enumerate}


\subsection{Step 3: Problem Solving}
\label{subsec:solving-task}
In the final step, the LLM uses the narrative generated in Step 2 to solve the original QA task. 
The structured and contextual understanding provided by the narrative supports LLM in accessing relevant aspects of the task.

%% file: latex/04-experimenal-setup.tex
\section{Experimental Setup}
\label{sec:experimental-setup}
To comprehensively evaluate the effectiveness of our proposed approach, we conduct experiments across a diverse set of tasks and models, employing various prompting techniques for comparison.

\subsection{Evaluation Tasks}
\label{subsec:tasks}

\input{latex/tables/gpqa_results}

We focus our evaluation on reasoning-intensive tasks spanning multiple domains, including physics, biology, and chemistry problem-solving. In particular, we utilize the \textbf{GPQA} (Diamond set) \cite{rein2024gpqa} and \textbf{JEEBench} \cite{arora-etal-2023-llms}.
GPQA is a Graduate-level Problem-solving QA dataset which comprises expert-crafted multiple-choice questions.
It contains 448 multiple-choice questions written by domain experts in biology, physics, and chemistry of high quality and difficulty.
We use the Diamond subset, which contains 198 questions on which both expert annotators agree.
JEEBench contains 515 challenging pre-engineering mathematics, physics and chemistry problems from the highly competitive IIT JEE-Advanced exam.

These problems span the different sciences and are extremely challenging, requiring in-depth reasoning and domain knowledge, making them well-suited for assessing our approach's ability to understand complex tasks and contextualize salient information within the problem space.

\subsection{Benchmarking Models}
\label{subsec:models}

To evaluate the performance of our approach across a wide range of Large Language Models, we experiment with the following LLM families:

\begin{enumerate}[leftmargin=*,noitemsep,topsep=0pt]
    \item \textbf{Meta}: Llama 3 8B \& Llama 3 70B
    \item \textbf{Mistral}: Mistral 7B \& Mixtral 8x7B
    \item \textbf{OpenAI}: GPT-3.5-turbo \& GPT 4
    \item \textbf{Microsoft}: Phi 3 Medium \& Phi 3 Mini
\end{enumerate}
These models were selected to cover a wide spectrum of capabilities, sizes and families, enabling a comprehensive evaluation of their strengths and limitations. 
More details on the implementation can be found in \autoref{appsec:model_details}.

\subsection{Methods Studied}
\label{subsec:approaches}

We compared our proposed approach against several prompting techniques:

\begin{itemize}[leftmargin=*,noitemsep,topsep=0pt]
    \item[] \textbf{Zero-shot Prompting}: This method, similar to our approach (SoT), does not rely on labeled examples. Instead, LLMs are prompted to solve tasks based solely on their pre-trained knowledge without any context provided. This approach serves as a baseline, demonstrating the LLMs' ability to solve problems without explicit guidance.
    
    \item[] \textbf{Zero-shot CoT}~\cite{wei2022chain}: This technique prompts the LLM to explicitly reason through the steps required to arrive at an answer. By prompting the model to generate a chain of thought, this method aims to improve the model's ability to solve complex problems by breaking them down into smaller, more manageable steps.

    \item[] \textbf{Tree of Thoughts}~\cite{yao2023tree}: This method systematically explores multiple reasoning paths instead of a single linear progression. In ToT, a tree-structured solution to a problem is generated by breaking it down into sub-problems. This approach enables the model to consider a broader set of potential solutions by evaluating each branch for correctness before proceeding further.

    \item[] \textbf{Graph of Thoughts}~\cite{besta2024got}: This technique extends the Tree of Thoughts (ToT) approach by allowing for a more flexible and non-hierarchical representation of problem-solving steps. In this method, the reasoning steps are treated as nodes, and the connections between them are edges that represent logical relationships or dependencies. In our experiments, we adopt the same setup described in their original work.

    \item[] \textbf{Analogical Reasoning}~\cite{yasunaga2023large}: This approach leverages analogies to help the model draw parallels between known concepts and the task at hand. By providing analogical examples, the model is guided to understand and apply similar reasoning patterns to new problems. In our experiment, we allow the LLMs to self-generate three exemplars for each question (akin to the prompt described in their paper). This enables them to identify relevant examples and adapt their reasoning accordingly.

    \item[] \textbf{Ours: Knowledge Identification}: To measure the effectiveness of our proposed approach, namely utilizing narrative in solving tasks, we prompt LLMs to solve the task based solely on the generated conceptual knowledge from Step 1 (described in Section \ref{subsec:question-clarification}). This allows us to compare the model capability in solving tasks using only the identified relevant knowledge versus leveraging this knowledge to structure a coherent narrative.
    
    \item[] \textbf{Ours: Story of Thought (SoT)}: This approach represents the core of our proposed method, where we leverage the generated narratives from Step 2 (described in Section \ref{subsec:narrative-generation}) to solve the given tasks.
    
\end{itemize}


%% file: latex/tables/gpqa_results.tex
\begin{table*}[tbhp!]
\centering
\resizebox{\textwidth}{!}{%
\begin{tabular}{lcccccccc}
\hline
\multicolumn{1}{c}{\multirow{2}{*}{Prompting Method}} & \multicolumn{2}{c}{\textbf{Meta}} & \multicolumn{2}{c}{\textbf{Mistral}} & \multicolumn{2}{c}{\textbf{OpenAI}} & \multicolumn{2}{c}{\textbf{Microsoft}} \\ \cline{2-9} 
\multicolumn{1}{c}{} & Llama 3 8B & Llama 3 70B & Mistral 7B & Mixtral 8x7B & ChatGPT 3.5 & GPT 4 & Phi-3 Mini & Phi-3 Medium \\ \hline
\textbf{Zero-shot} & 34.2 & 39.5 & 35.8 & 36.36 & 30.6 & 34.7 & \textbf{28.79} & 42.42 \\
\textbf{Zero-shot CoT} & 40.91 & 41.92 & 31.82 & 35.35 & 28.1 & 35.7 & 24.75 & 39.39 \\
\textbf{Tree of Thoughts} & 34.34 & 43.43 & 29.79 & 32.82 & 24.24 & 42.42 & 18.68 & 31.81 \\
\textbf{Graph of Thoughts} & 33.83 & 43.43 & 28.78 & 30.30 & 23.23 & 40.90 & 19.69 & 28.78 \\
\textbf{Analogical Reasoning (3-shot)} & 40.91 & 47.47 & 37.9 & 26.26 & 28.1 & 41.41 & 16.67 & \textbf{48.48} \\ \hline
\textbf{Ours: Knowledge Identification} & 40.4 & 48.99 & 35.35 & 37.77 & 27.77 & 40.90 & 20.71 & 37.88 \\
\textbf{Ours: Story of Thought (SoT)} & \textbf{43.43} & \textbf{51.01} & \textbf{38.4} & \textbf{38.89} & \textbf{30.8} & \textbf{48.98} & 22.73 & 36.36 \\ \hline
\end{tabular}%
}
\caption{On GPQA (Diamond set), Story of Thought (SoT) consistently outperforms other techniques. We present the performance (QA accuracy) of different methods with various LLMs on GPQA Diamond set.}
\label{tab:main-results-option-1}
\end{table*}

%% file: latex/05-results.tex
\section{Results}
\label{sec:results}

Our proposed SoT approach that incorporates narrative structures improves over almost all previous prompting approaches across two different problem-solving datasets.
This highlights the potential of using narratives to improve the ability of LLMs to understand and reason about the given information in various intensive reasoning tasks.

\input{latex/tables/jeebench_results}

\subsection{Performance on GPQA}
\label{subsec:main-results}

We present the results of our experiments on GPQA (Diamond) are presented in Table \ref{tab:main-results-option-1}. 
For this task, SoT is the best method to use with six of eight models.
The open-source Llama 3 70B model records the highest accuracy using the SoT method, achieving a score of 51.01\%. This is the highest accuracy observed among all models and methods tested in the study. Furthermore, the GPT-4 model shows the most notable improvement in accuracy when the SoT method is employed, compared to its zero-shot baseline. Specifically, the accuracy for GPT-4 increased from 34.7\% under zero-shot conditions to 48.98\% with SoT (i.e., an absolute increase of 14.28\%, or a relative increase of 41\% respectively).\footnote{We also find that Llama 3 70B with SoT outperforms zero-shot o1-preview which uses CoT style reasoning internally. \url{https://openai.com/index/learning-to-reason-with-llms/}.}

Interestingly, all reasoning strategies lead to an accuracy drop for the comparably smaller Phi-3 Mini model, and all CoT strategies except Analogical Reasoning also lead to the accuracy drop of the Phi-3 Medium model compared to its zero-shot baseline. We hypothesize that this is due to the low quality of the generated explanations (whether CoT steps or SoT narrative), as further studied in \S\ref{subsec:generating-narrative}.  

\begin{figure*}[!tbh]
\centering
\includegraphics[width=\textwidth]{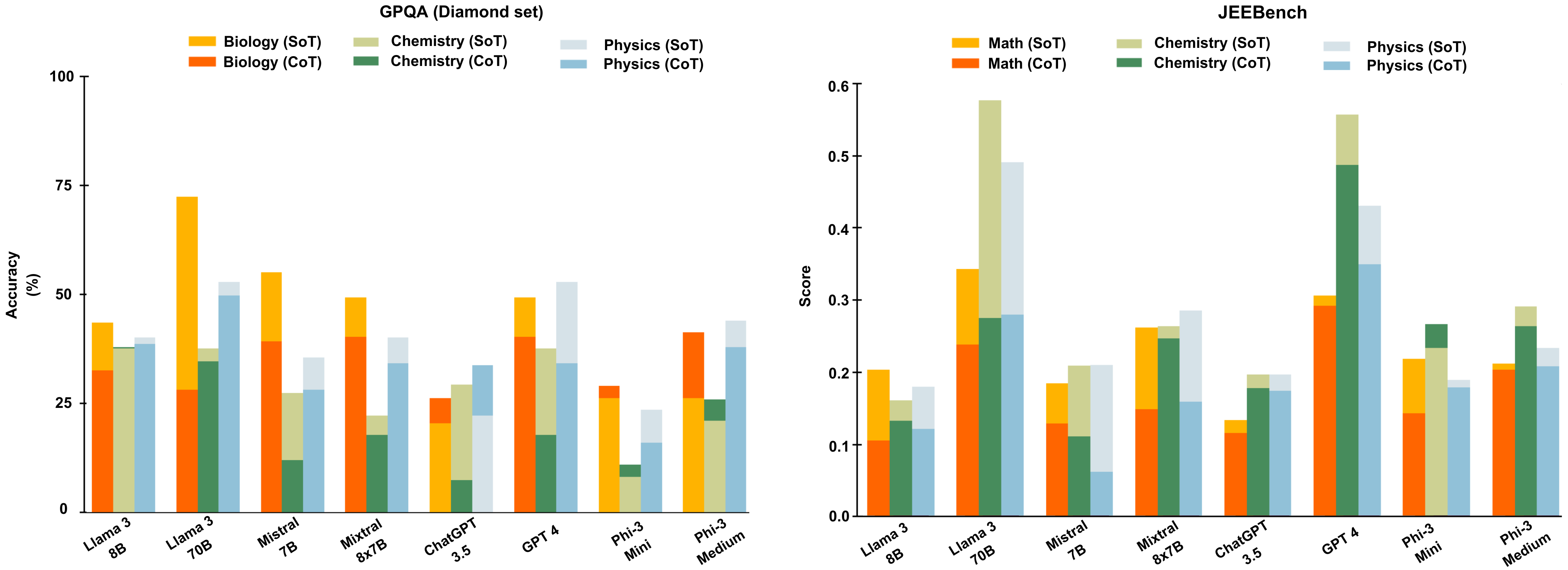}
\caption{Performance of \textbf{Story of Thought} (SoT) on GPQA and JEEBench across various LLMs and domains.}
\label{fig:domain}
\end{figure*}

\autoref{fig:domain} presents the performance of different models when using SoT across the different problem domains in GPQA.
We note that, on average, models improve the most on Biology problems when using SoT.
We hypothesize that this is because it is easier to simplify information for graduate level biology problems that can be used by models to come up with a solution.

\subsection{Performance on JEEBench}

\autoref{tab:performance-comparison} presents detailed experimental results on JEEBench.
Our proposed Story of Thought (SoT) method consistently improves the performance of seven out of the eight LLMs.
Using SoT, Llama 3 70B performance surpasses even the GPT models.
It obtains the highest scores in all subjects and question types (Except Single-Correct), with an overall aggregate score of 0.453.
This is a significant improvement on the previous SOTA, which was a strong GPT4 model used with both CoT and Self-Consistency.

Across models, the results highlight the effectiveness of Story of Thought (SoT) in enhancing model performance on complex, multi-disciplinary benchmarks like JEEBench, setting new SOTA results in several categories.
The improvements are particularly notable in the subject categories and question types where the other methods struggle.

In \autoref{fig:domain}, we present subject-wise performance of different models on JEEBench.
On average, model performance is highest on Chemistry problems when using SoT.
This is in contrast to findings on GPQA and could occur due to the difference in degree of difficulty of problems in the two datasets (graduate level vs high school level).
Regardless, improvements on Biology problems are not far behind those for Chemistry.

%% file: latex/tables/jeebench_results.tex
\begin{table*}[tbhp!]
\centering
\resizebox{0.96\textwidth}{!}{%
\begin{tabular}{l|ccc|cccc|c}
\hline
 & \textbf{Chemistry} & \textbf{Mathematics} & \textbf{Physics} & \textbf{Integer} & \textbf{Single-Correct} & \textbf{Multi-Correct} & \textbf{Numeric} & \textbf{Total} \\
\hline
GPT-4+CoT+SC@8* & 0.463 & 0.308 & 0.449 & 0.293 & \textbf{\textcolor{bluecolor}{{0.618}}} & 0.410 & 0.234 & 0.389 \\
\hline
Llama 3 8B & 0.143 & 0.082 & 0.089 & 0.061	& 0.127 & 0.148 & 0.044 & 0.102 \\
Llama 3 8B+CoT & 0.127 & 0.101 & 0.116 & \textbf{0.11} & 0.145 & 0.149 & 0.036 & 0.112 \\
Ours: Llama 3 8B+SoT &  \textbf{0.154} &  \textbf{0.195} &  \textbf{0.172} &  0.072 &  \textbf{0.259} & \textbf{0.324} & 0.028 &  \textbf{0.173} \\
\hline
Llama 3 70B & 0.324 & 0.189 & 0.274 & 0.171 & 0.345 & 0.316 & 0.131 & 0.25 \\
Llama 3 70B+CoT & 0.264 & 0.228 & 0.268 & 0.159 & 0.291 & 0.317 & 0.175 & 0.249 \\
Ours: Llama 3 70B+SoT & \textbf{\textcolor{bluecolor}{{0.554}}} & \textbf{\textcolor{bluecolor}{{0.329}}} & \textbf{\textcolor{bluecolor}{{0.471}}} & \textbf{\textcolor{bluecolor}{{0.446}}} & \textbf{0.42} & \textbf{\textcolor{bluecolor}{{0.485}}} & \textbf{\textcolor{bluecolor}{{0.462}}} & \textbf{\textcolor{bluecolor}{{0.453}}} \\
\hline
Mistral 7B & 0.119 & 0.079 & 0.091 & 0.049 & 0.109 & 0.159 & 0.022 & 0.094 \\
Mistral 7B+CoT & 0.106 & 0.123 & 0.059 & 0.073 & 0.118 & 0.165 & 0.022 & 0.102 \\
Ours: Mistral 7B+SoT & \textbf{0.2} & \textbf{0.177} & \textbf{0.201} & \textbf{0.11} & \textbf{0.245} & \textbf{0.224} & \textbf{0.146} & \textbf{0.19} \\
\hline
Mixtral 8x7B & 0.22	& 0.151	& 0.167 & 0.122 & 0.218 & 0.261 & 0.058 & 0.176 \\
Mixtral 8x7B+CoT & 0.237 & 0.142 & 0.152 & 0.061 & 0.209 & 0.27 & 0.08 & 0.173 \\
Ours: Mixtral 8x7B+SoT & \textbf{0.253} & \textbf{0.251} & \textbf{0.274} & \textbf{0.268} & \textbf{0.309} & 0.277 & \textbf{0.182} & \textbf{0.257} \\
\hline
ChatGPT 3.5 & \textbf{0.228} & \textbf{0.146} & 0.173 & 0.073 & \textbf{0.318} & \textbf{0.249} & 0.029 & \textbf{0.177} \\
ChatGPT 3.5+CoT & 0.17 & 0.111 & 0.167 & 0.11 & 0.173 & 0.206 & 0.051 & 0.142 \\
Ours: ChatGPT 3.5+SoT & 0.189 & 0.128 & \textbf{0.189} & 0.073 & 0.291 & 0.204 & 0.051 & 0.161 \\
\hline
GPT 4 & 0.423 & 0.212 & 0.352 & 0.207 & 0.455 & 0.383 & 0.153 & 0.309 \\
GPT 4+CoT & 0.468 & 0.280 & 0.335 & 0.256 & \textbf{0.473} & 0.448 & 0.175 & 0.350 \\
Ours: GPT 4+SoT & \textbf{0.535} & \textbf{0.294} & \textbf{0.413} & \textbf{0.378} & 0.4 & \textbf{0.453} & \textbf{0.321} & \textbf{0.395} \\
\hline
Phi-3 Mini & \textbf{0.256} & 0.12 & \textbf{0.199} & 0.146 & 0.255 & 0.224 & 0.08 & 0.18 \\
Phi-3 Mini+CoT & \textbf{0.256} & 0.137 & 0.171 &  0.134 & 0.209 & 0.216 & \textbf{0.139} & 0.181 \\
Ours: Phi-3 Mini+SoT & 0.224 & \textbf{0.209} & 0.181 & \textbf{0.183} & \textbf{0.282} & \textbf{0.234} & 0.124 & \textbf{0.207} \\
\hline
Phi-3 Medium & 0.298 & 0.193 & 0.165 & 0.146 & 0.255 & 0.286 & 0.139 & 0.218 \\
Phi-3 Medium+CoT & 0.253 & 0.195 & 0.199 & 0.171 & 0.236 & 0.274 & 0.139 & 0.214 \\
Ours: Phi-3 Medium+SoT & \textbf{0.279} & \textbf{0.203} & \textbf{0.224} & \textbf{0.232} & \textbf{0.273} & \textbf{0.263} & \textbf{0.153} & \textbf{0.231} \\
\hline
\end{tabular}%
}
\caption{On JEEBench, Story of Thought (SoT) outperforms previous SOTA as well as other methods. We present the aggregate score by subject as well as question type and present the overall aggregate score. The best overall scores are highlighted in \textbf{\textcolor{bluecolor}{{blue}}} while the best score by method for a model is in \textbf{bold}. \small * reported in \cite{arora-etal-2023-llms}.
}
\label{tab:performance-comparison}
\end{table*}

%% file: latex/05a-analysis.tex
\section{Analysis of SoT Aspects}


\subsection{Role of the Narrative Quality/Choice}
\label{subsec:generating-narrative}

The choice of \textit{narrator} model (i.e., the model that generates narratives) can impact the problem-solving resuls. In the following experiments, we apply the narratives generated by other large and small open-source LLMs to the Phi-3 Mini and Phi-3 Medium models. The results of these experiments are presented in Table~\ref{tab:role-of-narrative}.

\begin{table}[!tbh]
\centering
\small%
\begin{tabular}{lll}
\hline
\multirow{2}{*}{\textbf{Narrative Generator}} & \multicolumn{2}{c}{\textbf{Solver Models}} \\ \cline{2-3} 
 & \multicolumn{1}{c}{Phi-3 Mini} & \multicolumn{1}{c}{Phi-3 Medium} \\ \hline
Llama 3 8B & \multicolumn{1}{c}{23.74 (+1.01$\uparrow$)} & \multicolumn{1}{c}{37.88 (+1.28$\uparrow$)} \\
Llama 3 70B & \multicolumn{1}{c}{25.25 (+2.52$\uparrow$)} & \multicolumn{1}{c}{\textbf{39.39} (+2.79$\uparrow$)} \\
Mistral 7B & \multicolumn{1}{c}{24.24 (+1.51$\uparrow$)} & \multicolumn{1}{c}{38.38 (+1.78$\uparrow$)} \\
Mixtral 8x7B & \multicolumn{1}{c}{24.74 (+2.01$\uparrow$)} & \multicolumn{1}{c}{35.86 (-0.74$\downarrow$)} \\ \hline
\end{tabular}
\caption{Applying generated narratives by open-source models to Microsoft models to solve the tasks.}
\label{tab:role-of-narrative}
\end{table}

\input{latex/tables/narrative_technique_occurrence}

We observe that the \textbf{narratives} generated by most models 
\textbf{consistently improve the accuracy} of both Microsoft models compared to the baseline (i.e., when both models use their own generated narratives in Step 2 to solve the tasks, shown in Table~\ref{tab:main-results-option-1}). The absolute improvements range from 1.0\% to 2.8\%, with the Llama 3 70B model generating the most effective narratives. A slight decrease in accuracy is observed with the mixture-of-experts Mixtral 8x7B narratives for the Phi-3 Medium model, highlighting the need for careful selection and evaluation of narrator models to ensure compatibility and optimal performance.
Larger models generate narratives that break down problems to make them more easily solvable.
Unsurprisingly, there is larger room for improving the problem solving abilities of smaller models.



\subsection{Impact of Narrative Elements}
\label{subsec:narrative-techniques}

To measure the impact of each of the narrative techniques we jointly prompted on the performance of open-source Meta models, we ablate the designed prompt in Step 2 (of Section~\ref{subsec:narrative-generation}) to apply each of the techniques separately. The results in Table~\ref{tab:narrative-technique-accuracy} indicate that \textbf{employing any single narrative technique at a time is notably less effective at boosting QA accuracy than utilizing a combination of these simultaneously}.

\begin{table}[!ht]
\centering
\small%
\begin{tabular}{lcc}
\hline
\multicolumn{1}{c}{\multirow{2}{*}{Narrative Technique}} & \multicolumn{2}{c}{\textbf{Meta}} \\ \cline{2-3} 
\multicolumn{1}{c}{} & Llama 3 8B & Llama 3 70B \\ \hline

\textbf{Progressive Disclosure} & 34.85 (-8.58$\downarrow$) & 44.95 (-6.06$\downarrow$)  \\
\textbf{Branching} & 34.34 (\textbf{-9.09$\downarrow$}) & 44.95 (-6.06$\downarrow$)  \\
\textbf{Analogy} & 39.39 (-4.04$\downarrow$) & 46.46 (-4.55$\downarrow$)  \\
\textbf{Analogical Reasoning} & 40.4 (-3.03$\downarrow$) & 45.45 (-5.56$\downarrow$)  \\
\textbf{Metaphor} & 41.41 (-2.02$\downarrow$) & 44.44 (-6.57$\downarrow$)  \\ \hline
\textbf{All} & 43.43  & 51.01  \\ \hline
\end{tabular}%
\caption{Comparing accuracy when using a single narrative technique. The values in parentheses represent the decrease in accuracy percentage points compared to a combination of multiple narrative techniques simultaneously (shown in Table \ref{tab:main-results-option-1}).}
\label{tab:narrative-technique-accuracy}
\end{table}



For both models (Llama 3 8B and 70B), the decrease in accuracy is comparably smaller (-3.0\% to -5.6\%) when using only the analogical components of the narrative (\textit{Analogy} and \textit{Analogical Reasoning}) than when using only the structural instructions (\textit{Progressive Disclosure} or \textit{Branching}) which leads to larger (-6.0\% to -9.1\%) accuracy loss. 
However, reasoning alone does not perform on par with the full narrative generation listing all the techniques. Prompting for \textit{Metaphor} usage only leads to a larger accuracy loss in the 70B model (-6.6\%) compared to the smaller one (-2.0\%). 
This makes it difficult to determine both how the narrative techniques relate to each other and whether the model truly comprehends the prompts it receives.
We disentangle and study the two going forward.

\subsection{Analyzing Generated Narratives}
\label{subsec:analyzing-generated-narratives}

\input{latex/tables/similarity_metrics}

To gain deeper insights into the generated narratives, we designed a prompt (shown below) that utilizes our best-performing model (Llama 3 70B) to annotate the number of occurrences of each narrative technique for each generated narrative by all models used in our experiments. We can better interpret how the model executed the narrative technique prompt, by asking it to label if and where the mentioned techniques are used in the text generated. Less frequently labeled techniques might be the ones where LLM doesn't have a clear understanding of what it is asked to do. 
A proportion of the techniques and their correlation can provide us with a better picture of LLM's interpretation of the instruction as well.
We detail the instruction given to the LLM in \autoref{appsec:prompts}.

\begin{figure}[!t]
\centering
\includegraphics[width=\columnwidth]{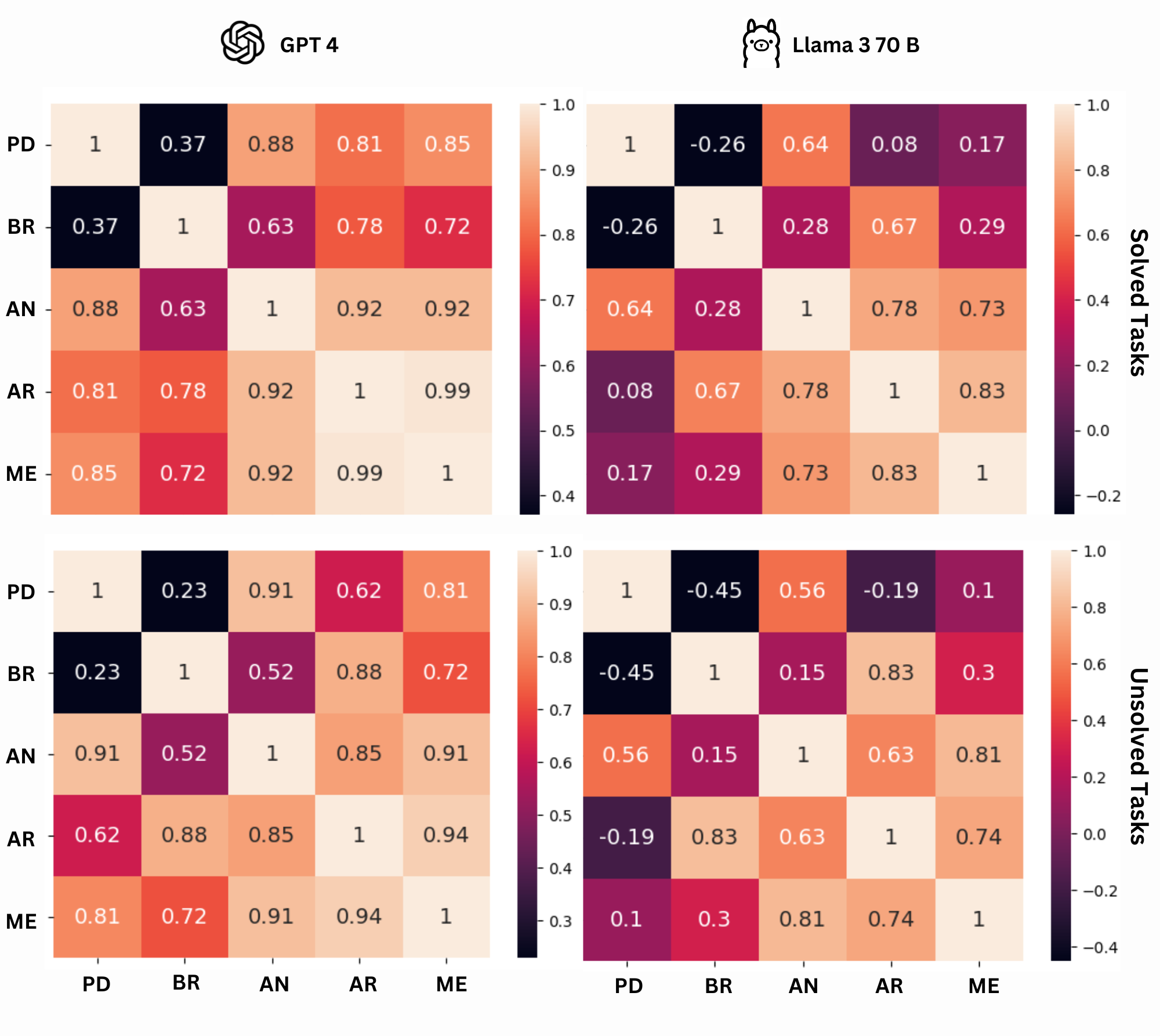}
\caption{Correlation coefficients among all narrative techniques (\textbf{PD} = Progressive Disclosure, \textbf{BR} = Branching, \textbf{AN} = Analogy, \textbf{AR} = Analogical Reasoning, \textbf{ME} = Metaphor) used in the SoT approach for GPT 4 and Llama 3 70B in solved and unsolved tasks.}
\label{correlation}
\end{figure}

We aim to uncover patterns and variations in the use of narrative techniques across different LLMs. \autoref{tab:narrative-technique-Occurrence} compares the total number of occurrences for each narrative technique across various LLMs.
\paragraph{Variability in Utilization of Narrative Techniques Across Models:} In our designed prompt in Step 2 (i.e., Narrative Generation), LLMs generate narrative using all 5 narrative techniques. However, as Table \ref{tab:narrative-technique-Occurrence} indicates, not all techniques were employed equally. This reveals that while some techniques like \textit{Analogy} and \textit{Progressive Disclosure} were consistently utilized, others such as \textit{Branching} were applied less frequently.

We observe a trend across all LLM families where models with larger capacities, such as Llama 3 70B and GPT-4, consistently show higher occurrences of narrative techniques compared to their smaller counterparts. Furthermore, OpenAI's models (ChatGPT 3.5 \& GPT-4) demonstrate the highest total occurrences of narrative techniques, with 2,338 and 2,097, respectively with a notable emphasis on \textit{Metaphors} and \textit{Analogies}.

\paragraph{Correlation Among Narrative Techniques:}
To further investigate the dynamics of narrative techniques, we compute correlations between the frequencies of narrative techniques across solved and unsolved tasks, as shown in Figure \ref{correlation}. This analysis aims to uncover if the models consistently use certain narrative techniques together or vary significantly. Our initial results indicate diverse correlation patterns, suggesting that the effectiveness of narrative techniques in solving tasks across various LLMs needs to be further analyzed.

\subsection{Analyzing SoT Reasoning}
\label{subsec:analyzing-sot-reasoning}

Table \ref{tab:similarity} compares the similarity of Story of Thought (SoT) and Chain of Thought (CoT) reasoning outputs to human explanations for different language models on the GPQA (Diamond set) dataset, using BertScore, ROUGE-L, and BLEU.

The differences between ROUGE-L values are insignificant and do not display any clear trends.
However, according to BLEU scores, using SoT results in explanations closer to humans and the differences are more pronounced.

As per BertScore (an embedding-based similarity metric), Llama 3 models (8B and 70B) explanations are more similar to human ones when using SoT reasoning across all three metrics.
However, Mistral models (7B and 8x7B), GPT-4, and Phi-3 Mini generate explanations more similar to human explanations when using CoT reasoning across all metrics.
The semantic similarity of narratives generated by Llama 3 70B to human explanations combined with their effect of improving smaller models indicates that these narratives present information about the problems in a simplified manner.

%% file: latex/tables/narrative_technique_occurrence.tex
\begin{table*}[!tbh]
\centering
\resizebox{\textwidth}{!}{%
\begin{tabular}{lcccccccc}
\hline
\multicolumn{1}{c}{\multirow{2}{*}{Narrative Technique}} & \multicolumn{2}{c}{\textbf{Meta}} & \multicolumn{2}{c}{\textbf{Mistral}} & \multicolumn{2}{c}{\textbf{OpenAI}} & \multicolumn{2}{c}{\textbf{Microsoft}} \\ \cline{2-9} 
\multicolumn{1}{c}{} & Llama 3 8B & Llama 3 70B & Mistral 7B & Mixtral 8x7B & ChatGPT 3.5 & GPT 4 & Phi-3 Mini & Phi-3 Medium \\ \hline
\textbf{Progressive Disclosure} & 427 & 597 & 191 & 191 & 744 & 570 & 367 & 368 \\
\textbf{Branching} & 30 & 56 & 51 & 20 & 72 & 168 & 34 & 61 \\
\textbf{Analogy} & 418 & 425 & 117 & 161 & 498 & 595 & 569 & 499 \\
\textbf{Analogical Reasoning} & 205 & 191 & 78 & 108 & 213 & 336 & 276 & 206 \\
\textbf{Metaphor} & 249 & 316 & 103 & 137 & 811 & 428 & 418 & 291 \\ \hline
$\sum$ & 1329 & 1585 & 540 & 617 & 2338 & 2097 & 1664 & 1425 \\ \hline
\end{tabular}%
}
\caption{Comparing Generated Narratives - Total Number of Occurrences for each
Narrative Techniques (Evaluator: Llama 3 70B)}
\label{tab:narrative-technique-Occurrence}
\end{table*}

%% file: latex/tables/similarity_metrics.tex
\begin{table*}[!tbh]
\centering
\small%
\resizebox{\textwidth}{!}{%
\begin{tabular}{lcccccc}
\hline
\multicolumn{1}{c}{\multirow{2}{*}{Similarity Metric}} & \multicolumn{2}{c}{\textbf{BertScore}} & \multicolumn{2}{c}{\textbf{ROUGE-L}} & \multicolumn{2}{c}{\textbf{BLEU}} \\ \cline{2-7} 
\multicolumn{1}{c}{} & SoT Reasoning & CoT Reasoning & SoT Reasoning & CoT Reasoning & SoT Reasoning & CoT Reasoning \\ \hline
\textbf{Llama 3 8B} & \textbf{0.28} & 0.06 & \textbf{0.19} & 0.11 & \textbf{6.57} & 0.19 \\
\textbf{Llama 3 70B} & \textbf{0.3} & 0.04 & \textbf{0.2}	& 0.1 & \textbf{8.18} & 0.06 \\ \hline
\textbf{Mistral 7B} & 0.27 & \textbf{0.33} & 0.18 & \textbf{0.2} & \textbf{8.12} & 4.65 \\
\textbf{Mixtral 8x7B} & 0.3	& \textbf{0.34} & 0.19 & \textbf{0.21} & \textbf{8.92} & 8.14 \\ \hline
\textbf{ChatGPT 3.5} & \textbf{0.3} & 0.24 & \textbf{0.19} & 0.16 & 6.1 & \textbf{6.07} \\
\textbf{GPT 4} & 0.31 & \textbf{0.34} & 0.19 & \textbf{0.2} & \textbf{8.84} & 6.73 \\ \hline
\textbf{Phi-3 Mini} & 0.27 & \textbf{0.31} & 0.17 & \textbf{0.19} & \textbf{6.54} & 6.36 \\
\textbf{Phi-3 Medium} & 0.3 & \textbf{0.35} & 0.2 & \textbf{0.21} & 7.13 & \textbf{8.4} \\ \hline
\end{tabular}%
}
\caption{Comparison of generated Story of Thought (SoT) and Chain of Thought (CoT) reasoning with Human Explanations on the GPQA (Diamond set) using BERTScore, ROUGE-L, and BLEU metrics across various large language models. Bold values indicate the reasoning approach that is more similar to human explanations for each model and metric pair.}
\label{tab:similarity}
\end{table*}

%% file: latex/07-conclusion.tex
\section{Conclusion} 
\label{sec:conclusion}

Inspired by findings from human cognitive processes explored in didactics research, in this work, we propose to use narrative techniques in LLM prompting.
We present strong evidence on public benchmark datasets that narrative techniques have the potential to notably enhance the reasoning abilities of  LLMs in complex problem-solving tasks. By incorporating narrative structures, which mimic human cognitive processes of organizing and interpreting information, LLMs can achieve higher levels of performance and provide more contextually enriched responses.




%% file: latex/08-limitations.tex
\section*{Limitations}
\label{sec:limitations}

\paragraph{Contribution limitations.} 
The occurrences of narrative techniques do not necessarily imply the quality or effectiveness of the generated narratives; rather, they provide insights into the models' tendencies and preferences in employing these techniques. Therefore, answering the question of why narrative is helping LLMs is more complex and needs to be further investigated by looking into different research areas such as cognitive and communication theories.


\paragraph{Dataset limitations.} So far, we used only GPQA and JEEBench problems as the most challenging set of problem-solving benchmarks we were aware of. Other comparable benchmarks, such as MGSM, are much closer to human or superhuman accuracy already without reasoning prompts and will be explored in future work. 

\paragraph{Analysis limitations.} We used Llama 70 B to respectively analyze the narratives. The intuition behind this experiment is that we can better interpret how the model executed the narrative technique prompt, by asking it to label if and where the mentioned techniques are used in the text generated. An alternative would be a thorough human assessment and further analysis of the impact on downstream performance, both of which we pursue in ongoing follow-up experiments.
(We also previously prompted the LLMs in Step 2 to explain each of these five narrative techniques to make sure the concepts are understood before generating the narrative.)

%% file: latex/09-appendix.tex
\clearpage

\appendix

\begin{table*}[tbh]
\centering
\resizebox{\textwidth}{!}{%
\begin{tabular}{lcccccc}
\hline
\multicolumn{1}{c}{\multirow{2}{*}{Prompting Method}} & \multicolumn{2}{c}{\textbf{Meta}} & \multicolumn{2}{c}{\textbf{Mistral}} & \multicolumn{2}{c}{\textbf{Microsoft}} \\ \cline{2-7} 
\multicolumn{1}{c}{} & Llama 3 8B & Llama 3 70B & Mistral 7B & Mixtral 8x7B & Phi-3 Mini & Phi-3 Medium \\ \hline
\textbf{Zero-shot} & 30.81 (-3.39$\downarrow$) & 31.31 (-8.19$\downarrow$) & 19.7 (-16.1$\downarrow$) & 18.18 (-18.18$\downarrow$) & 29.8 (+1.01$\uparrow$) & 21.72 (-20.7$\downarrow$) \\
\textbf{Zero-shot CoT} & 27.27 (-13.64$\downarrow$) & 33.33 (-8.59$\downarrow$) & \textbf{22.73} (-9.09$\downarrow$) & 17.17 (-18.18$\downarrow$) & 32.32 (+7.57$\uparrow$) & 21.21 (-18.18$\downarrow$) \\
\textbf{Analogical Reasoning} & 27.78 (-13.13$\downarrow$) & 40.91 (-6.56$\downarrow$) & 10.61 (-27.29$\downarrow$) & 19.19 (-7.07$\downarrow$) & \textbf{35.86} (+19.19$\uparrow$) & 16.67 (-31.81$\downarrow$) \\ \hline
\textbf{Ours: Knowledge Identification} & 32.32 (-8.08$\downarrow$) & 42.4 (-6.59$\downarrow$) & 16.67 (-18.68$\downarrow$) & 14.65 (-23.12$\downarrow$) & 28.28 (+7.57$\uparrow$) & 23.26 (-14.62$\downarrow$) \\
\textbf{Ours: Story of Thought (SoT)} & \textbf{34.85} (-8.58$\downarrow$) & \textbf{45.4} (-5.61$\downarrow$) & 20.2 (-18.2$\downarrow$) & \textbf{20.2} (-18.69$\downarrow$) & 27.7 (+4.97$\uparrow$) & \textbf{25.75} (-10.85$\downarrow$) \\ \hline
\end{tabular}%
}
\caption{Performance of various LLMs across different prompting methods on GPQA (Diamond set). Correct answers are presented in the second option. Values in parentheses indicate the change in accuracy compared to the original setting in Table \ref{tab:main-results-option-1} where the correct answer was in the first option.}
\label{tab:main-results-option-2}
\end{table*}

\section{Robustness of LLM Predictions}
\label{appsec:correct-option}
In the original GPQA dataset used for our experiments, the correct answers are always presented as the first option among the multiple choices. However, To further evaluate the robustness of the LLMs, we conduct an additional experiment where the correct answers are placed in the second option instead. Table~\ref{tab:main-results-option-2} presents the results of these experiments, comparing the performance of various prompting methods across six different open-source LLMs.
We observe that most LLMs experience a significant drop in accuracy when the correct answer is moved to the second option. However, despite the overall decrease in accuracy, our proposed approach, Story of Thought (SoT), consistently outperforms the baseline methods for most LLMs. The SoT method achieves the highest accuracy for the Meta Llama 3 8B, Meta Llama 3 70B, Mistral 8x7B, and Microsoft Phi-3 Medium models, demonstrating its effectiveness in enhancing the robustness of LLMs to changes in the problem structure.

\section{Model Implementation Details}
\label{appsec:model_details}

All experiments, except for those involving OpenAI models, were conducted on local machines equipped with GPUs. The models were run locally on a GPU setup without quantization using the Hugging Face Transformer library\footnote{\url{https://huggingface.co/docs/transformers}}. For OpenAI's GPT-3.5-turbo and GPT 4 models, we use the OpenAI API to generate outputs. Across all models, we use a temperature of 1.0 and a maximum number of tokens of 8,000 and report the accuracy.

\section{Prompts Used in Story of Thought}
\label{appsec:prompts}

We describe the prompts used for each stage in pipeline.

\subsection{Question Clarification}

\lstset{
  basicstyle=\ttfamily\small,
  columns=fullflexible,
  frame=single,
  breaklines=true,
  linewidth=0.47\textwidth
}

\begin{lstlisting}
You are an explorer who wants to identify and collect different related and specialized subject areas to clarify the question. Your goal is to narrow down the question and provide relevant areas of knowledge and experience you have that help clarify the question mentioned below. You should not answer the question.

<question>
\end{lstlisting}

\subsection{Narrative Generation}

\lstset{
  basicstyle=\ttfamily\small,
  columns=fullflexible,
  frame=single,
  breaklines=true,
  linewidth=0.47\textwidth
}

\begin{lstlisting}
You are an expert in narrative-based explanations for science communication. Your goal is to clarify the following question in a narrative way through the interconnected information provided below to enable a non-expert to comprehend the question in a more coherent and contextually rich manner. You should not answer the question.

Make sure to use all of these narrative techniques when clarifying the question through the interconnected information: Progressive Disclosure, Branching, Analogy, Analogical Reasoning, and Metaphor.

<question>

<generated information in the previous step>
\end{lstlisting}

\subsection{Problem Solving}

\lstset{
  basicstyle=\ttfamily\small,
  columns=fullflexible,
  frame=single,
  breaklines=true,
  linewidth=0.47\textwidth
}

\begin{lstlisting}
You are an expert in analyzing narrative-based explanations for solving tasks. Please answer the following question based on the following narrative-based clarification:

<question>

Options:
<options>

<generated narrative in the previous step>
\end{lstlisting}

\subsection{Analyzing Generated Narratives}

\lstset{
  basicstyle=\ttfamily\small,
  columns=fullflexible,
  frame=single,
  breaklines=true,
  linewidth=0.47\textwidth
}

\begin{lstlisting}
You are an expert in analyzing narrative-based explanations for science communication. Your goal is to find out which narrative techniques have been used in the following narrative-based explanation.

Label the narrative-based explanation using the following narrative-based techniques:
1. Progressive Disclosure
2. Branching
3. Analogy
4. Analogical Reasoning
5. Metaphor

<generated narrative>
\end{lstlisting}

\section{Performance on JEEBench}
\label{appsec:jeebench-full-performance}

\begin{table*}[htbp!]
\centering
\small%
\resizebox{\textwidth}{!}{%
\begin{tabular}{l|ccc|cccc|c}
\hline
 & \textbf{Chemistry} & \textbf{Mathematics} & \textbf{Physics} & \textbf{Integer} & \textbf{Single-Correct} & \textbf{Multi-Correct} & \textbf{Numeric} & \textbf{Total} \\
\hline
GPT-4+CoT+SC@8* & 0.463 & 0.308 & 0.449 & 0.293 & \textbf{\underline{0.618}} & 0.410 & 0.234 & 0.389 \\
\hline
Llama 3 8B & 0.143 & 0.082 & 0.089 & 0.061	& 0.127 & 0.148 & 0.044 & 0.102 \\
Llama 3 8B+CoT & 0.127 & 0.101 & 0.116 & \textbf{0.11} & 0.145 & 0.149 & 0.036 & 0.112 \\
Llama 3 8B+Analogical Reasoning (3-shot) & 0.139 & 0.111 & 0.128 & \textbf{0.11} & 0.145 & 0.165 & \textbf{0.058} & 0.124 \\
Ours: Llama 3 8B+Knowledge Identification & 0.199 & 0.099 & 0.134 & 0.073 & 0.227 & 0.171 & 0.058 & 0.137 \\
Ours: Llama 3 8B+SoT &  \textbf{0.154} &  \textbf{0.195} &  \textbf{0.172} &  0.072 &  \textbf{0.259} & \textbf{0.324} & 0.028 &  \textbf{0.173} \\
\hline
Llama 3 70B & 0.324 & 0.189 & 0.274 & 0.171 & 0.345 & 0.316 & 0.131 & 0.25 \\
Llama 3 70B+CoT & 0.264 & 0.228 & 0.268 & 0.159 & 0.291 & 0.317 & 0.175 & 0.249 \\
Llama 3 70B+Analogical Reasoning (3-shot) &  0.314	& 0.24 & 0.295 & 0.195 & 0.318 & 0.349 & 0.19 & 0.276 \\
Ours: Llama 3 70B+Knowledge Identification & 0.317 & 0.226 & 0.254 & 0.195 & 0.345 & 0.323 & 0.146 & 0.26 \\
Ours: Llama 3 70B+SoT & \textbf{\underline{0.554}} & \textbf{\underline{0.329}} & \textbf{\underline{0.471}} & \textbf{\underline{0.446}} & \textbf{0.42} & \textbf{\underline{0.485}} & \textbf{\underline{0.462}} & \textbf{\underline{0.453}} \\
\hline
Mistral 7B & 0.119 & 0.079 & 0.091 & 0.049 & 0.109 & 0.159 & 0.022 & 0.094 \\
Mistral 7B+CoT & 0.106 & 0.123 & 0.059 & 0.073 & 0.118 & 0.165 & 0.022 & 0.102 \\
Mistral 7B+Analogical Reasoning (3-shot) & 0.157 & 0.084 & 0.116 & 0.073 & 0.155 & 0.169 & 0.029 & 0.114 \\
Ours: Mistral 7B+Knowledge Identification & 0.109 & 0.055 & 0.063 & 0.037 & 0.091 & 0.117 & 0.022 & 0.073 \\
Ours: Mistral 7B+SoT & \textbf{0.2} & \textbf{0.177} & \textbf{0.201} & \textbf{0.11} & \textbf{0.245} & \textbf{0.224} & \textbf{0.146} & \textbf{0.19} \\
\hline
Mixtral 8x7B & 0.22	& 0.151	& 0.167 & 0.122 & 0.218 & 0.261 & 0.058 & 0.176 \\
Mixtral 8x7B+CoT & 0.237 & 0.142 & 0.152 & 0.061 & 0.209 & 0.27 & 0.08 & 0.173 \\
Mixtral 8x7B+Analogical Reasoning (3-shot) & 0.202 & 0.155 & 0.197 & 0.122 & 0.191 & \textbf{0.281} & 0.066 & 0.179 \\
Ours: Mixtral 8x7B+Knowledge Identification & 0.184 & 0.129 & 0.144  & 0.122 & 0.155 & 0.237 & 0.044 & 0.15 \\
Ours: Mixtral 8x7B+SoT & \textbf{0.253} & \textbf{0.251} & \textbf{0.274} & \textbf{0.268} & \textbf{0.309} & 0.277 & \textbf{0.182} & \textbf{0.257} \\
\hline
ChatGPT 3.5 & \textbf{0.228} & \textbf{0.146} & 0.173 & 0.073 & \textbf{0.318} & \textbf{0.249} & 0.029 & \textbf{0.177} \\
ChatGPT 3.5+CoT & 0.17 & 0.111 & 0.167 & 0.11 & 0.173 & 0.206 & 0.051 & 0.142 \\
ChatGPT 3.5+Analogical Reasoning (3-shot) & 0.208 & 0.125 & 0.148 & 0.098 & 0.2 & 0.216 & \textbf{0.073} & 0.156\\
Ours: ChatGPT 3.5+Knowledge Identification & 0.155 & 0.141 & 0.167 & \textbf{0.122} & 0.209 & 0.188 & \textbf{0.073} & 0.151 \\
Ours: ChatGPT 3.5+SoT & 0.189 & 0.128 & \textbf{0.189} & 0.073 & 0.291 & 0.204 & 0.051 & 0.161 \\
\hline
GPT 4 & 0.423 & 0.212 & 0.352 & 0.207 & 0.455 & 0.383 & 0.153 & 0.309 \\
GPT 4+CoT & 0.468 & 0.280 & 0.335 & 0.256 & \textbf{0.473} & 0.448 & 0.175 & 0.350 \\
GPT 4+Analogical Reasoning (3-shot) & 0.479 & 0.286 & 0.396 & 0.305 & 0.4 & 0.43 & 0.307 & 0.371 \\
Ours: GPT 4+Knowledge Identification & 0.481 & 0.287 & 0.386 & 0.293 & 0.373 & 0.452 & 0.314 & 0.373 \\
Ours: GPT 4+SoT & \textbf{0.535} & \textbf{0.294} & \textbf{0.413} & \textbf{0.378} & 0.4 & \textbf{0.453} & \textbf{0.321} & \textbf{0.395} \\
\hline
Phi-3 Mini & \textbf{0.256} & 0.12 & \textbf{0.199} & 0.146 & 0.255 & 0.224 & 0.08 & 0.18 \\
Phi-3 Mini+CoT & \textbf{0.256} & 0.137 & 0.171 &  0.134 & 0.209 & 0.216 & \textbf{0.139} & 0.181 \\
Phi-3 Mini+Analogical Reasoning (3-shot) & 0.205 & 0.159 & 0.195 & 0.146 & 0.264 & 0.218 & 0.088 & 0.182 \\
Ours: Phi-3 Mini+Knowledge Identification & 0.168 & 0.091 & 0.106 & 0.073 & 0.136 & 0.181 & 0.044 & 0.118 \\
Ours: Phi-3 Mini+SoT & 0.224 & \textbf{0.209} & 0.181 & \textbf{0.183} & \textbf{0.282} & \textbf{0.234} & 0.124 & \textbf{0.207} \\
\hline
Phi-3 Medium & 0.298 & 0.193 & 0.165 & 0.146 & 0.255 & 0.286 & 0.139 & 0.218 \\
Phi-3 Medium+CoT & 0.253 & 0.195 & 0.199 & 0.171 & 0.236 & 0.274 & 0.139 & 0.214 \\
Phi-3 Medium+Analogical Reasoning (3-shot) & 0.258 & 0.181 & 0.173 & 0.159 & 0.218 & 0.276 & 0.117 & 0.202 \\
Ours: Phi-3 Medium+Knowledge Identification & 0.288 & 0.163 & 0.205 & 0.207 & 0.236 & 0.235 & 0.161 & 0.211 \\
Ours: Phi-3 Medium+SoT & \textbf{0.279} & \textbf{0.203} & \textbf{0.224} & \textbf{0.232} & \textbf{0.273} & \textbf{0.263} & \textbf{0.153} & \textbf{0.231} \\
\hline
\end{tabular}%
}
\caption{On JEEBench, Story of Thought (SoT) outperforms previous SOTA as well as other methods. We present the aggregate score by subject as well as question type and present the overall aggregate score. * denotes SOTA results taken from the original paper \cite{arora-etal-2023-llms}.
}
\label{tab:performance-comparison-jeebench}
\end{table*}

\section{Story of Thought (SoT) vs. Chain of Thought (CoT)}
\label{sot-cot-examples}

\begin{figure*}[hbt!]
\centering
\includegraphics[width=\textwidth]{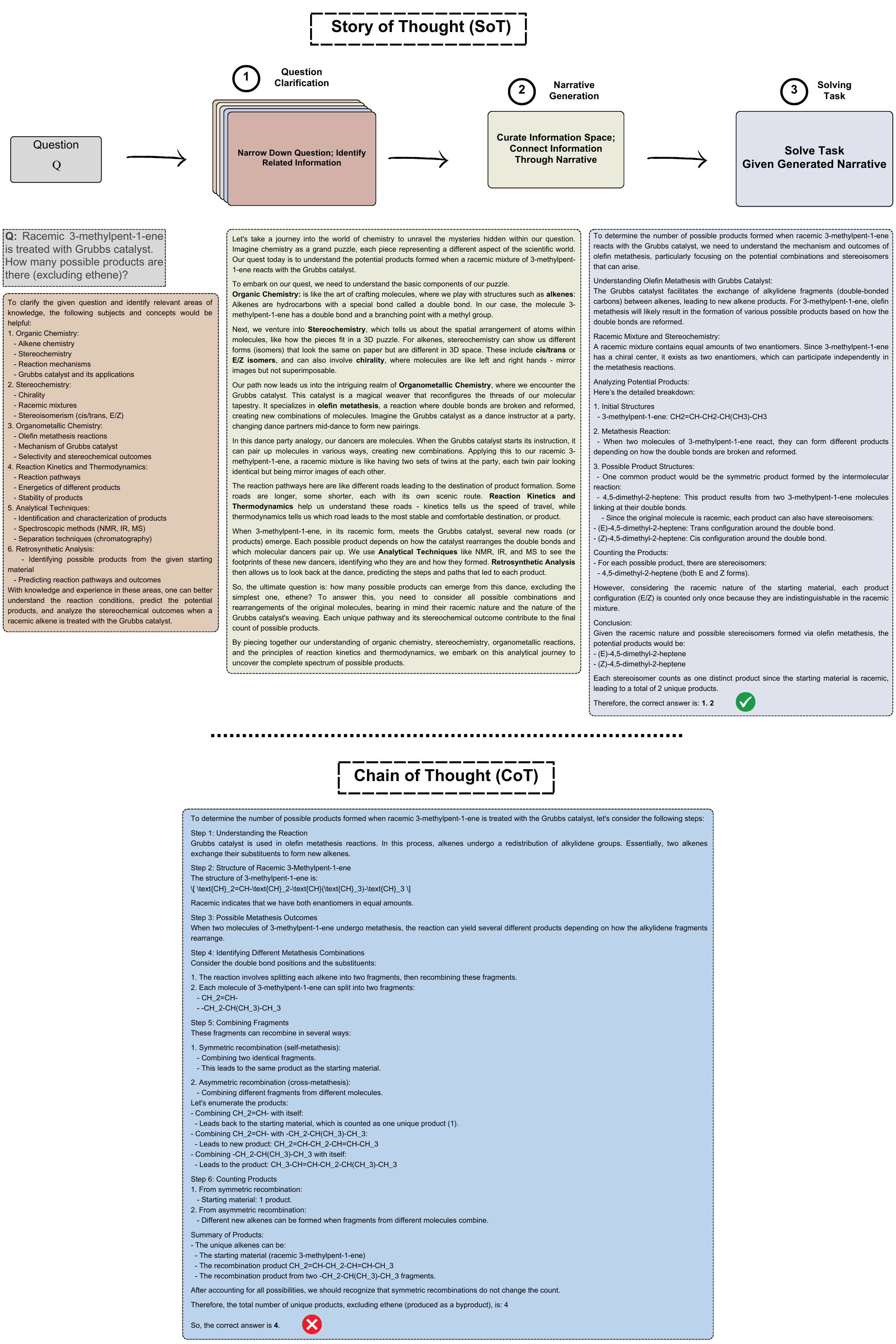}
\caption{An actual example of SoT.}
\end{figure*}